\documentclass[10pt,conference]{IEEEtran}
\IEEEoverridecommandlockouts
\usepackage{amsmath,graphicx}
\usepackage{amsmath, amsfonts, amssymb,graphicx}
\usepackage{booktabs} 
\usepackage{multirow} 
\usepackage{cite} 
\usepackage{tabularx}
\usepackage{stackengine}
\usepackage{array}
\usepackage{xcolor}
 \usepackage{threeparttable}
 \usepackage{url}

\definecolor{myred}{RGB}{0,0,0}
\newcommand{\Sisdr}{SDR}
\usepackage{color, colortbl}
\definecolor{Gray}{gray}{0.9}
\usepackage{cite}
\usepackage{amsmath,amssymb,amsfonts}
\usepackage{algorithmic}
\usepackage{graphicx}
\usepackage{textcomp}
\usepackage{xcolor}
\def\BibTeX{{\rm B\kern-.05em{\sc i\kern-.025em b}\kern-.08em
    T\kern-.1667em\lower.7ex\hbox{E}\kern-.125emX}}
\begin{document}

\title{Joint Beamforming and Speaker-Attributed ASR \\for Real Distant-Microphone Meeting Transcription
}

\author{
\IEEEauthorblockN{Can Cui\thanks{This work was mostly conducted while Can Cui was pursuing her PhD with Inria Centre at Université de Lorraine (Nancy, France).}}
\IEEEauthorblockA{\textit{iFLYTEK Co., Ltd.} \\
Shanghai, China \\
cancui11@iflytek.com
}
\and
\IEEEauthorblockN{Imran Sheikh}
\IEEEauthorblockA{\textit{Vivoka} \\
Metz, France \\
imran.sheikh@vivoka.com
}
\and
\IEEEauthorblockN{Mostafa Sadeghi, Emmanuel Vincent}
\IEEEauthorblockA{\textit{Université de Lorraine, CNRS, Inria, LORIA, F-54000 } \\
Nancy, France \\
\{mostafa.sadeghi, emmanuel.vincent\}@inria.fr
}

}

\maketitle

\begin{abstract}
Distant-microphone meeting transcription is a challenging task. State-of-the-art end-to-end speaker-attributed automatic speech recognition (SA-ASR) architectures lack a multichannel noise and reverberation reduction front-end, which limits their performance. In this paper, we introduce a joint beamforming and SA-ASR approach for real meeting transcription. We first describe a data alignment and augmentation method to pretrain a neural beamformer on real meeting data. We then compare fixed, hybrid, and fully neural beamformers as front-ends to the SA-ASR model. Finally, we jointly optimize the fully neural beamformer and the SA-ASR model. Experiments on the real AMI corpus show that,
while state-of-the-art method based on channel fusion fails to improve ASR performance, fine-tuning SA-ASR on the fixed beamformer's output and jointly fine-tuning SA-ASR with the neural beamformer reduce the word error rate by 8\% and 9\% relative, respectively.
\end{abstract}


\section{Introduction}
\label{sec:intro}

Transcription of real distant-microphone conversational meetings or domestic data is an active research area \cite{watanabe2020chime,yu2022m2met,cornell2023chime}. It remains challenging today due to noise, reverberation, and overlapping speech. 
To improve performance, many studies have employed a front-end multichannel speech separation module or a series of (fixed, statistical, or neural) beamformers steered toward the speakers to extract individual speech signals from the overlapping speech mixture and subsequently feed each of them to a single-speaker ASR module \cite{yoshioka2018recognizing,kanda2019guided}. \textcolor{myred}{Unfortunately, the separation error then propagates to the ASR module.}
Later studies 
\cite{chang2019mimo,wu2021investigation,shi2022train} 
proposed jointly optimizing ASR and front-end separation by back-propagating ASR losses to the separation module via permutation invariant training (PIT), albeit at higher computational cost.

End-to-end multi-speaker ASR based on serialized output training (SOT) \cite{kanda2020serialized} addresses these limitations. \cite{kanda2021end} introduced an end-to-end Transformer-based speaker-attributed ASR (SA-ASR) system for joint recognition of speech and speaker identities from single-channel log Mel features, later extended to multichannel SA-ASR (MC-SA-ASR) by integrating log Mel \cite{yu2023mfcca} and phase \cite{cui2023} features with time-varying multi-frame cross-channel attention (MFCCA). Such multichannel attention schemes are often believed to outperform classical beamforming yielding state-of-the-art performance. Yet,
in contrast to a frequency-dependent complex-valued beamformer, they rely on frequency-independent real-valued weights, which results in limited noise and reverberation reduction.

In this paper, we propose to combine SA-ASR with a beamforming-based noise and reverberation reduction front-end to improve speech and speaker recognition in far-field conditions. The beamformer fuses the mixture channels into a single-channel enhanced mixture fed to SA-ASR. While such a front-end is common in single-speaker scenarios, extending it to real multi-speaker scenarios is non trivial. First, 
the beamformer must dynamically adapt its spatial response based on the speakers' positions and activity patterns, which vary over time. This complexity explains the scarcity of multi-speaker beamforming front-ends in the literature.
BeamformIt \cite{anguera2007acoustic} was used as a front-end for PIT-based multi-speaker ASR \cite{chang2019mimo} and single-speaker ASR baselines for multi-speaker ASR tasks \cite{ochiai2017unified,watanabe2020chime}, while minimum variance distortion-less response (MVDR) beamforming was used as a front-end for SA-ASR in \cite{kanda2023vararray} without comparison to MFCCA. To our knowledge, no full-neural beamforming front-end has been used for SA-ASR so far. Second, pretraining a neural beamformer on real meeting data is challenging due to the absence of ground truth noiseless dry mixture signals as pretraining targets. 

The contributions of this paper are as follows. First, we introduce data alignment and augmentation techniques to pretrain a multi-speaker neural beamformer on a real meeting corpus containing both distant microphone and headset recordings. \textcolor{myred}{Note that the beamformer is employed to reduce noise and reverberation, but not to extract the individual speakers}. Second, we propose a pipeline integrating beamforming with SA-ASR, aiming to improve both speech and speaker recognition. Third, we evaluate the differences in performance between statistical, \textcolor{myred}{hybrid, }and neural beamformers. Finally, we jointly optimize the neural beamformer and the SA-ASR model. Our experiments on the AMI corpus \cite{carletta2005ami} reveal that, 
while MFCCA-based channel fusion does not improve ASR performance, fine-tuning SA-ASR on the fixed beamformer's output and jointly fine-tuning SA-ASR with the neural beamformer reduces the WER by 8\% and 9\% relative, respectively.

The paper is organized as follows. Section \ref{sec:related} reviews the beamformers considered in this work and SA-ASR model. Section~\ref{sec:methods} introduces our joint system and the AMI data alignment and augmentation pipeline. Section \ref{sec:exp} describes our experimental setup and results, and we conclude in Section \ref{sec:conclusion}.

\section{Background}
\label{sec:related}

\subsection{Beamforming and dereverberation methods}
\label{subsec:das}

The delay-and-sum (DAS) beamformer \cite{don1993array} is a fixed beamformer, which depends only on the delays between the microphone signals and a reference microphone. 
It involves computing the delays using a time difference of arrival estimator such as the generalized cross-correlation with phase transform \cite{knapp1976generalized}, shifting the phase of the microphone signals accordingly in the complex short-time Fourier transform domain, and summing them.

\textcolor{myred}{Deep neural network (DNN)-based MVDR beamforming \cite{lu2022towards,kim23b_interspeech} combines neural networks with traditional beamforming methods. The DNN is trained to estimate masks in the time-frequency domain that enhance the desired signals and suppress interference. This information is then used to compute the MVDR beamformer weights, which minimize the output power while preserving the signals from the target direction. }
This method can be seen as a transition between traditional optimization-based beamforming and fully neural network-based approaches.

The Filter-and-Sum Network (FaSNet) system \cite{luo2019fasnet} aims to directly estimate time-domain beamforming filters. 
It employs a two-stage design: the first stage estimates filters for a reference microphone, and the second stage estimates filters for the remaining microphones based on pairwise cross-channel features between the pre-separated output and each microphone.
The FaSNet architecture utilizes dual-path recurrent neural networks \cite{luo2020dual} to extract information from both the channel and frame levels. 
The Transform-average-concatenate (TAC) \cite{luo2020end} design paradigm addresses channel permutation and is capable of handling various numbers of microphones.

\textcolor{myred}{Multichannel dereverberation using weighted prediction error (WPE) \cite{nakatani2010speech} reduces reverberation by modeling and subtracting late reverberant components from the observed audio signal using linear prediction. This technique optimizes prediction coefficients and error weights to enhance speech clarity and intelligibility in reverberant environments. }

\subsection{Speaker-attributed ASR}
\label{subsec:saasr}


A Transformer-based end-to-end SA-ASR system was proposed in \cite{kanda2021end}. Following the SOT principle \cite{kanda2020serialized}, the output is the concatenation of all speakers' sentences in first-in-first-out order, where each token is associated with one speaker ID and distinct speakers are separated by an \textless sc\textgreater\ token. The inputs to the model consist of an acoustic feature sequence (log-Mel filterbank) and a set of reference speaker embeddings. A Conformer-based ASR Encoder first encodes acoustic information, along with a speaker encoder to encode speaker information. Then, the Transformer-based ASR decoder and speaker decoder modules decode text and speaker information, respectively. 
The speaker decoder generates a speaker representation corresponding to each token in the ASR Decoder's output token sequence. This representation is used to assign speakers by computing a dot product with the reference speaker embeddings.

\section{Proposed methods}
\label{sec:methods}

\subsection{Real meeting data alignment and augmentation for neural beamformer training}
\label{subsec:data-aug}

\begin{figure}[!htbp]
  \vspace{-3pt}
  \centering
  \includegraphics[width=0.75\linewidth, height=4cm]{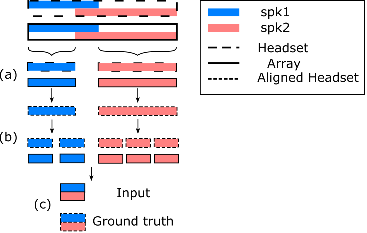}
  \vspace{-7pt}
  \caption{Mixture generation from real meeting data.}
  \label{fig:real}
  \vspace{-8pt}
\end{figure}

\textcolor{myred}{Neural beamforming on real-world far-field data is challenging due to the lack of ground truth enhanced signals for training. Real meeting corpora such as AMI include headset recordings for each speaker, but these can't be used directly as ground truth because of variable delays caused by the speakers' positions. To address this, we generate aligned array and headset signals in three steps (see Fig.~\ref{fig:real}): (a) extract all non-overlapping speech segments for each speaker based on dataset annotations; (b) align each non-overlapping headset segment with the corresponding array segment using matched filters, then cut them into fixed-length clips; (c) randomly sample and mix array clips from different speakers to create far-field mixtures, and mix the corresponding aligned headset clips to obtain the reference noise and reverberation-free mixtures enhanced mixtures.}


The matched filters $f_{ij}(t)$ in step (b) are calculated in the least squares sense by solving
\begin{equation}\label{eq:eq1}
\begin{aligned}
\min_{f_{ij}}\sum_{t} (f_{ij}\star h_j(t) - x_i(t))^2
\end{aligned}
\end{equation}
where $h_j(t)$ and $x_i(t)$ stand for the headset signal of speaker $j$ and the array signal at microphone $i$, respectively, and $\star$ denotes time-domain convolution. The solution is obtained as the finite impulse response Wiener filter,
\textcolor{myred}{which is commonly employed for filter estimation.}

 
\subsection{Joint multichannel beamforming and SA-ASR}
\begin{figure}[!tbp]
  \centering
  \includegraphics[width=0.8\linewidth]{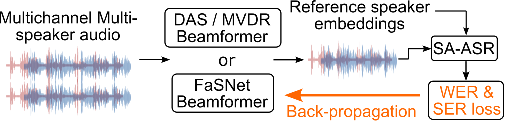}
  \vspace{-7pt}
  \caption{Proposed joint system of Beamformer and SA-ASR}
  \vspace{-5pt}
  \label{fig:joint-system}
   \vspace{-5pt}
\end{figure}

We propose a joint system integrating beamforming and SA-ASR for multichannel, distant-microphone meeting transcription. As illustrated in Fig.~\ref{fig:joint-system}, the multichannel audio is first processed by a beamformer to generate enhanced single-channel audio. The output audio is then fed to SA-ASR to obtain speech and speaker recognition results. We compare the performance of the fixed DAS beamformer, \textcolor{myred}{the hybrid DNN-MVDR (simply refered to as MVDR) beamformer,} and the fully neural FaSNet beamformer, when fine-tuning the SA-ASR model on training data enhanced with the respective beamformer. In addition, we backpropagate the loss from SA-ASR to FaSNet, in order to fine-tune the neural beamformer according to the SA-ASR training objective.

\section{Experimental evaluation}
\label{sec:exp}
This section details our experimental setup and results. For reproducibility, our code is available online.\footnote{\url{https://github.com/can-cui/joint-beamforming-sa-asr}}
\subsection{Datasets}
\label{subsec:data}

\noindent \textbf{Mixed AMI} --- To train the \textcolor{myred}{MVDR and the} FaSNet beamformer, we apply the method described in Section~\ref{subsec:data-aug} to the AMI meeting corpus. This method creates mixtures of real single-speaker AMI segments and their corresponding ground truths. We name this dataset Mixed AMI. We only use \textcolor{myred}{one-quarter} of all meetings
and fix the clip length to 4~s. The mixtures are generated by overlapping randomly selected clips from 1 to 4 speakers. The training, development, and test sets contain 150~h, 17~h, and 16~h of speech data, respectively.\\[.75em]
\textbf{Multi-speaker LibriSpeech} --- To optimize performance on AMI, the SA-ASR model requires pretraining on a larger simulated dataset \cite{yang2023simulating}. We created a 960~h training set and a 20~h development set from LibriSpeech \cite{panayotov2015LibriSpeech} train-960 and dev-clean.
We adopted the room simulation and speaker mixture settings described in \cite{cui2023}. 
\\[.75em]
\textbf{Real AMI} --- Once pretrained on Multi-speaker LibriSpeech, the SA-ASR model is fine-tuned and evaluated on real AMI multiple distant microphones (MDM) data. We utilize the segmentation method in \cite{cui2023} to partition the MDM data into 5~s chunks and adjust the chunk start/end times to non-overlapped word boundaries.
The resulting Real AMI dataset contains respectively 165~h, 19~h, and 19~h for training, development, and test. For both Mixed AMI and Real AMI, we consider 2- and 8-channel settings. 

\subsection{Metrics}
\label{subsec:metrics}
We evaluate the beamforming performance using the scale-invariant signal-to-distortion ratio (SI-SDR) and its improvement (SI-SDRi), implemented in Asteroid toolkit \cite{Pariente2020Asteroid}, measured in dB on the Mixed AMI test set. \textcolor{myred}{We calculate the baseline SI-SDR for SA-ASR by defining the array mixture signal as the estimated signal, ensuring that the SI-SDRi is 0~dB without beamforming. For all beamforming methods, we calculate the SI-SDR by defining the beamformed signal as the estimated signal and compute SI-SDRi by subtracting the corresponding  baseline SI-SDR}. The performance of SA-ASR is evaluated by the word error rate (WER) and the sentence-level speaker error rate (SER) \cite{kanda2020joint} in \%.  
\subsection{Models description}
\label{subsec:systems}
We utilize the implementation of DAS from the SpeechBrain toolkit \cite{speechbrain}. \textcolor{myred}{For the MVDR model, we utilize the implementation in TorchAudio \cite{yang2021torchaudio}.
 The number of filterbank output channels and the number of bins in the estimated masks are both 513.} The implementation of FaSNet with TAC is from the Asteroid toolkit.
 The encoder dimension and feature dimension are 64. The dual path blocks consist of a 4-layer dual model. 

We implemented SA-ASR and the MFCCA-based MC-SA-ASR system in \cite{cui2023} as a baseline using SpeechBrain. In SA-ASR and MC-SA-ASR, the Conformer-based encoder, the Transformer-based decoder and the speaker decoder have 12, 6 and 2 layers, respectively. All multi-head attention mechanisms have 4 heads,  the model dimension is 256, and the size of the feedforward layer is 2,048. 
The speaker embedding model is a pretrained\footnote{Available at \url{https://huggingface.co/speechbrain/spkrec-ecapa-voxceleb}} Emphasized Channel Attention, Propagation, and Aggregation in Time-Delay Neural Network \cite{desplanques2020ecapa},
 yielding 192 dimensional embeddings. 

\textcolor{myred}{Additionally, we test the performance obtained with WPE, implemented by \cite{Drude2018NaraWPE}, during the evaluation of the MC-SA-ASR model. }

\subsection{Training setup}
\label{subsec:setup}
The \textcolor{myred}{MVDR and} FaSNet beamformer are trained on Mixed AMI for 200 epochs with early stopping, using the Adam optimizer with a learning rate of $10^{-3}$.

The ASR modules in SA-ASR and MC-SA-ASR are pretrained on Multi-speaker LibriSpeech for 80 epochs using Adam with a learning rate of $5\times 10^{-4}$. The ASR and speaker modules are then further pretrained on Multi-speaker LibriSpeech for 60 epochs with a learning rate of $2.5\times 10^{-4}$. 
SA-ASR and MC-SA-ASR are fine-tuned on either unprocessed (baseline) or beamformed Real AMI data for 15 epochs, using Adam with a learning rate of $3\times 10^{-4}$. 

\subsection{Evaluation results}

\newcolumntype{g}{>{\columncolor{Gray}}c}
\begin{table*}[t]
\caption{Results for models fine-tuned and tested on unprocessed (SA-ASR and MC-SA-ASR) or beamformed (DAS-SA-ASR, MVDR-SA-ASR, FaSNet-SA-ASR) data. For convenience, we denote SI-SDR and SI-SDRi as SDR and SDRi, respectively.}
\vspace{-10pt}
\label{table:beamformers}
\centering
\scalebox{0.85}{
\addtolength{\tabcolsep}{-0.4em}
\begin{tabular}{c|c|c|cgcgcgcg|cgcgcgcg}
    \toprule
    \multirow{4}{*}{\bfseries System} & 
    \multirow{4}{*}{\bfseries \# Prm} & 
    \multirow{4}{*}{\bfseries \# Chn} & 
    \multicolumn{8}{c|}{\bfseries Mixed AMI test set} & 
    \multicolumn{8}{c}{\bfseries Real AMI test set} 
    \\ 
    \cmidrule(lr){4-11} \cmidrule(lr){12-19} 
    &&& \multicolumn{2}{c}{\bfseries 1-spk} & \multicolumn{2}{c}{\bfseries2-spk } &\multicolumn{2}{c}{\bfseries3-spk }& \multicolumn{2}{c|}{\bfseries1,2,3,4-spk }&\multicolumn{2}{c}{\bfseries1-spk} & \multicolumn{2}{c}{\bfseries2-spk } &\multicolumn{2}{c}{\bfseries3-spk }& \multicolumn{2}{c}{\bfseries1,2,3,4-spk } \\
    \cmidrule(lr){4-5}  \cmidrule(lr){6-7}  \cmidrule(lr){8-9}  \cmidrule(lr){10-11}  \cmidrule(lr){12-13} \cmidrule(lr){14-15} \cmidrule(lr){16-17} \cmidrule(lr){18-19}
    &&&\textbf{\Sisdr}& \textbf{\Sisdr i}& \textbf{\Sisdr}& \textbf{\Sisdr i}&\textbf{\Sisdr}& \textbf{\Sisdr i}& \textbf{\Sisdr}&\textbf{\Sisdr i}&\textbf{WER}& \textbf{SER} & \textbf{WER} & \textbf{SER}& \textbf{WER}& \textbf{SER}& \textbf{WER}& \textbf{SER}\\ 
    \cmidrule(lr){1-19}
    SA-ASR &69M&{1}&5.41&0&5.75&0&5.79&0&5.66&0&26.76&11.92&40.23&32.66&52.31&45.11&44.54 &34.73 \\
    
    \hline
    {MC-SA-ASR} &\multirow{2}{*}{59M}&{2}&5.41&0&5.75&0&5.79&0&5.66&0&26.41&11.73&40.79&32.64&52.59&43.82&44.99 &34.43 \\
    \textcolor{myred}{+WPE (test)} &&{2}&5.72&0.31&5.93&0.18&5.92&0.13&5.87&0.21&26.43&12.13&40.80&32.54&52.32&44.14&44.72&34.65 \\
    \hline
\multirow{2}{*}{DAS-SA-ASR} &\multirow{3}{*}{69M}&\multirow{1}{*}{2} &5.62&0.21&5.42&-0.33&5.23&-0.56&5.39&-0.27&25.59&12.82&40.36&33.87&52.04&45.55&44.03&35.56\\
   &&\multirow{1}{*}{8}&5.66&0.25&5.48&-0.27&5.30&-0.49&5.38&-0.28&23.51&12.13&38.41&33.12&50.43&43.44&41.71 &34.40 \\
   \multirow{1}{*}{\textcolor{myred}{+WPE}}&&\multirow{1}{*}{8}&6.18&0.77&5.54&-0.21&5.04&-0.75&5.35&-0.31&23.49&11.43&37.89&32.67&50.12&43.59&41.37&\textbf{33.84} \\
   \hline
\multirow{2}{*}{MVDR-SA-ASR} &\multirow{3}{*}{74M}&\multirow{1}{*}{2} &7.40&1.98&7.42&1.67&7.46&1.80&7.44&1.78&26.54&12.94&41.07&34.47&52.81&45.18&44.39&35.92\\
   &&\multirow{1}{*}{8}&8.14&2.72&8.10&2.34&8.10&2.30&8.11&2.44&27.31&12.42&41.27&34.52&52.63&45.07&44.23&35.21\\
   \multirow{1}{*}{\textcolor{myred}{+WPE}}&&\multirow{1}{*}{8}& 8.40&2.99&8.09&2.34&8.03&2.24&8.14&2.48&26.35&12.93&41.03&33.72&52.12&44.26&44.12&35.37\\
   \hline

    
    \multirow{2}{*} {FaSNet-SA-ASR} &\multirow{4}{*} {72M}&{2}&10.21&4.79&9.85&4.09&9.56&3.76&\textbf{9.76}&\textbf{4.10}&26.86&11.33&40.91&35.67&52.57&47.12&44.57&36.24 \\
    &&{8}&10.41&4.99&10.01&4.25&9.72&3.92&\textbf{9.96}&\textbf{4.29}&26.53&10.73&39.93&34.78&51.70&45.28&44.11&35.51 \\
    
    \multirow{1}{*}{\textcolor{myred}{+WPE}}&&{8}&5.88&0.47&5.88&0.47&5.66&-0.13&5.85&0.19& 26.16&10.78&39.35&34.89&51.22&45.25&43.39&35.41\\

    \bottomrule
\end{tabular}
}
\begin{tablenotes}
\centering
      \footnotesize
      \item \textcolor{black}{Note: For all the tables, we employed the SCTK toolkit \cite{sctk} to conduct significance tests, specifically the Matched Pair Sentence Segment test. For the mixture of 1,2,3 and/or 4 speakers, the best WER/SER and the results statistically equivalent to it at a 0.05 significance level are highlighted.}
    \end{tablenotes}
\vspace{-15pt}
\end{table*}
 
\subsubsection{Fine-tuning SA-ASR with DAS vs.\ with frozen MVDR and FaSNet}
\label{subsec:eval1}

We initially evaluated an SA-ASR model fine-tuned on the first channel of Real AMI. The resulting WER and SER on mixtures of 1, 2, 3, or 4 speakers were 44.54\% and 34.73\%, respectively. However, when testing the same model on FaSNet beamformed 2-channel Real AMI, the WER, and SER increased to 64.32\% and 46.30\%. 
Therefore, in all following experiments, we fine-tune the SA-ASR model on real AMI training data enhanced using the same beamformer as the test data. This adaptation is essential to align the model with the specific conditions of the test set.

Table~\ref{table:beamformers} shows the test results of the baseline models (SA-ASR and MC-SA-ASR) and the combination of SA-ASR with three beamformers, where the parameters of MVDR and FaSNet are frozen during fine-tuning. \textcolor{myred}{Without WPE, }the WER comparison between SA-ASR (44.54\%) and MC-SA-ASR (44.99\%) demonstrates that, while MFCCA had achieved a 13\% relative WER reduction on simulated data in \cite{cui2023}, it is inefficient on real meeting data. In general, fine-tuning the SA-ASR model on beamformed audio improves the ASR performance, particularly in the 8-channel setting, where using the DAS beamformer leads to a {6\%} relative reduction in WER 
\textcolor{myred}{without WPE (41.71\%) and 8\% with WPE (40.96\%)} compared to SA-ASR. It is also interesting to note that, despite FaSNet's superior denoising and dereverberation performance in terms of {SI-SDRi}, the SA-ASR model trained on speech beamformed by FaSNet performs less effectively than the one trained on speech beamformed by DAS. In the 8-channel setting, \textcolor{myred}{without WPE,} using DAS results in a 5\% relative reduction in WER compared to using FaSNet (from 44.11\% to 41.71\%).
\textcolor{myred}{The WER relative reduction is up to 6\% (from 44.23\% to 41.71\%) compared to the MVDR-SA-ASR system. The latter system has a similar performance to the FaSNet-SA-ASR system.}

\begin{figure}[!t]
  \centering
  \includegraphics[width=0.85\linewidth, height=5.1cm]{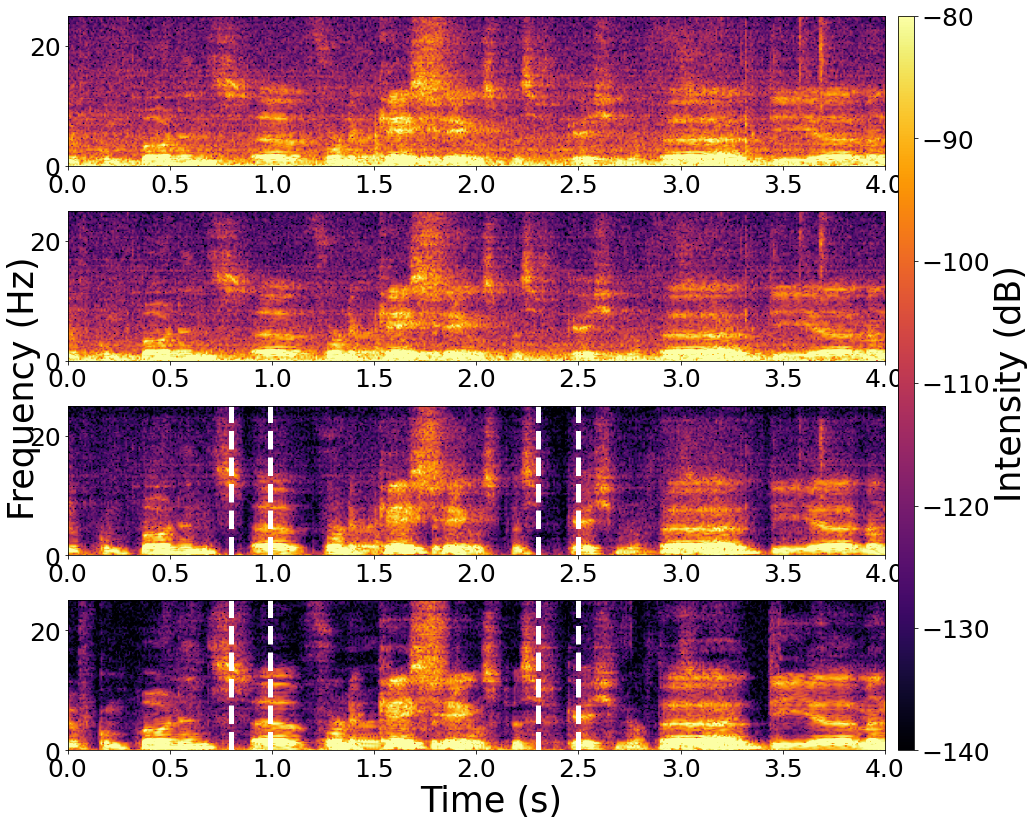}
  \vspace{-10pt}
  \caption{Spectrogram of one 8-channel Mixed AMI test chunk. From top to bottom: 1st array channel; DAS beamformed signal; FaSNet beamformed signal; ground truth.}
  \label{fig:beamformers}
  \vspace{-15pt}
\end{figure}

To find the reason for the difference between DAS-SA-ASR and FaSNet-SA-ASR, we visualize the spectrogram of one 8-channel Mixed AMI test chunk before and after beamforming (see Fig.~\ref{fig:beamformers}). It can be seen that, although FaSNet exhibits effective denoising, it also removes a portion of the speech signal, as highlighted by the white columns in the figure. On the contrary, DAS can preserve a significant portion of all speech signals while providing some denoising, which results in better speech and speaker recognition results.

\begin{table}[!tbp]
\caption{Results for jointly trained 2-channel FaSNet and SA-ASR models. ``Total": test set with 1,2,3,4-speaker mix.}
\vspace{-10pt}
\label{table:joint}
\centering
\scalebox{0.85}{
\addtolength{\tabcolsep}{-0.5em}
\begin{tabular}{ccccc|cccccc}
    \toprule
    \multicolumn{5}{c|}{\bfseries  Mixed AMI test set} & 
    \multicolumn{6}{c}{\bfseries Real AMI test set}
    \\ 
     \cmidrule(lr){1-5} \cmidrule(lr){6-11}
    \multicolumn{3}{c}{\bfseries Pretrained} & 
    \multicolumn{2}{c|}{\bfseries Fine-tuned} & 
    \multicolumn{2}{c}{\bfseries 1-spk} & 
    \multicolumn{2}{c}{\bfseries 3-spk mix}& 
    \multicolumn{2}{c}{\bfseries Total}
    \\ 
    \cmidrule(lr){1-3} \cmidrule(lr){4-5} \cmidrule(lr){6-7}  \cmidrule(lr){8-9} \cmidrule(lr){10-11} 
    \textbf{\# Epo}& \textbf{\Sisdr}& \textbf{\Sisdr i}& \textbf{\Sisdr}& \textbf{\Sisdr i}&  \textbf{WER}& \textbf{SER}& \textbf{WER}& \textbf{SER}& \textbf{WER}& \textbf{SER}\\ 
    \cmidrule(lr){1-11}
    0&5.66&0&-16.21&-21.87&25.31&11.37&48.63&43.07&41.71&\textbf{33.68}  \\
    5&9.27&3.61&5.13&-0.53&24.91 &13.06&47.49&45.00&\textbf{40.60}&34.87\\
    10&9.46&3.79&6.12&0.46&24.99&11.80&48.02&44.75&\textbf{40.82}&34.29\\
    
    50&9.69&4.02&7.05&1.39&24.54&13.51&47.71&43.87&\textbf{40.52}&34.43 \\
    
    \bottomrule
\end{tabular}
}
\vspace{-15pt}
\end{table}

Table~\ref{table:beamformers} also shows the performance differences for each system with or without WPE for dereverberation. First, even without beamforming, using WPE only during the inference phase for MC-SA-ASR results in a 0.21~dB improvement in SI-SDR and a slight absolute WER reduction of 0.27\% (from 44.99\% to 44.72\%). 
 For systems using beamformed signals, integrating WPE during the beamforming phase improves the SI-SDRi for DAS and MVDR but not for FaSNet. However, using WPE during beamforming to fine-tune the SA-ASR model systematically improves ASR and speaker identification performance. 
This demonstrates that WPE aids in reverberation reduction for fixed beamformers. However, the improvement in recognition performance for neural beamformers (MVDR and FaSNet) is less pronounced, likely because these beamformers have already learned to reduce reverberation during their training process.

\subsubsection{Joint optimization of FaSNet and SA-ASR}
\label{subsec:eval3}

Moreover, we pretrain FaSNet for 0, 5, 10, or 50 epochs and subsequently fine-tune it for 15 epochs jointly with SA-ASR by backpropagating the SA-ASR loss.
The results in Table~\ref{table:joint} show that joint optimization of FaSNet and SA-ASR (40.52\%) reduces the WER \textcolor{myred}{by 9\% relative to the frozen FaSNet (44.57\%) and to SA-ASR (44.54\%). We also observed the lowest SER as 33.68\%, 7\% relatively lower than using the frozen FaSNet (36.24\%).} However, the fine-tuned FaSNet exhibits a smaller SI-SDRi than the pretrained one. This indicates that joint training optimizes FaSNet for ASR performance rather than maximum noise and reverberation reduction at the cost of greater speech distortion. Furthermore, while the number of FaSNet pretraining epochs significantly impacts the SI-SDRi, it does not significantly impact the result of joint optimization, provided it's nonzero. This demonstrates the insensitivity of the joint optimization to the pretraining level. 


\section{Conclusion}
\label{sec:conclusion}
This paper explored the integration of beamforming with SA-ASR for joint speech and speaker recognition of far-field meeting audio. {We evaluated the impact of fine-tuning SA-ASR on the outputs of \textcolor{myred}{DAS, MVDR, or FaSNet} beamformers and jointly fine-tuning SA-ASR with the latter, and compared it with state-of-the-art MFCCA-based channel fusion. Experiments revealed that, in contrast to previously published results on simulated data, MFCCA's performance is limited on real AMI data. This highlights the importance of systematically evaluating SA-ASR on real meeting data.  Experiments show that DAS and jointly trained FaSNet-SA-ASR reduce WER by 8\% and 9\%, respectively, while adding WPE to DAS-SA-ASR yields a 3\% gain in both WER and SER.} 

\bibliographystyle{IEEEbib}
\bibliography{strings}

\end{document}